# Advanced deep architecture pruning using single filter performance

Yarden Tzach[1,+], Yuval Meir[1,+], Ronit D. Gross[1,+], Ofek Tevet[1], Ella Koresh[1] and Ido Kanter[1,2,*]

[1]Department of Physics, Bar-Ilan University, Ramat-Gan, 52900, Israel.

[2]Gonda Interdisciplinary Brain Research Center, Bar-Ilan University, Ramat-Gan, 52900, Israel.

[+] These authors contributed equally

[*]Corresponding author email: ido.kanter@biu.ac.il

**ABSTRACT**. Pruning the parameters and structure of neural networks reduces the computational complexity, energy consumption, and latency during inference. Recently, a novel underlying mechanism for successful deep learning (DL) was presented based on a method that quantitatively measures the single filter performance in each layer of a DL architecture, and a new comprehensive mechanism of how deep learning works was presented. This statistical mechanics inspired viewpoint enables to reveal the macroscopic behavior of the entire network from the microscopic performance of each filter and their cooperative behavior. Herein, we demonstrate how this understanding paves the path to high quenched dilution of the convolutional layers of deep architectures without affecting their overall accuracy using applied filter's cluster connections (AFCC). AFCC is exemplified on VGG-11 and EfficientNet-B0 architectures trained on CIFAR-100, and its high pruning outperforms other techniques using the same pruning magnitude. Additionally, this technique is broadened to single nodal performance and highly pruning of fully connected layers, suggesting a possible implementation to considerably reduce the complexity of over-parameterized AI tasks.

I. INTRODUCTION

Deep learning (DL) architectures are commonly used to perform complex computational tasks such as object classification[1-3]. To achieve a high performance in the computational tasks, the architectures use hundreds of convolutional layers (CLs)[4-6]. As classification tasks become more complex, deeper architectures are required to achieve enhanced accuracies[7,8], although they are known to be heavily over-parameterized[9]. This growth in the size and parameters of convolutional neural networks (CNNs) inevitably results in high latency and energy consumption, and they may be incompatible with their deployment on mobile or embedded devices, rather than more compact networks[10].

Pruning is a common method for reducing network complexity, where a fraction of the connections and nodes of the trained network are permanently removed. Such a quenched dilution is mostly performed with additional fine tuning without affecting the overall accuracies[11]. Many quenched pruning methods rely on the correlation between the weight strength and its importance to the accuracy of the network[12]. However, these techniques operate on each individual weight rather than as an overall method; thus, they cannot achieve the utmost pruning or large hefty pruning without affecting the overall accuracy. Another technique involves pruning entire layers [13,14]. However, this approach is limited by the number of layers that can be pruned before the accuracy is diminished[15].

Recently, a quantitative method to explain the underlying mechanism of successful DL[16,17] was presented, enabling the quantification of the importance and functionality of each filter in a CNN architecture. This mechanism indicates that each filter mainly recognizes a small subset of labels from the entire dataset, known as the filter's cluster, and in addition generates small noise on the successive layer from elements outside its cluster. The mechanism is based on several steps. For a trained CNN [Fig. 1, Stage 1], the weights of the first $m$ trained layers remain unchanged, and their outputs are fully connected (FC) with random initial weights assigned to the output layer, which represent the labels [Fig. 1, Stage 2]. The output of the first $m$ layers represent the preprocessing of an input using the partial deep architecture, and the FC layer is trained to minimize loss, which is a relatively simple computational task.

The trained weights of the FC layer are used in the final step to quantify the functionality of each filter [Fig. 1, Stage 3]. The single-filter performance is determined by silencing all the weights of the FC layer except for the specific weights that emerge from a single filter. At this point, the train inputs are presented and preprocessed by the first $m$ layers while influencing the output units only through the small aperture of one filter. The results demonstrate that each filter essentially identifies a small subset of output labels organized into a small cluster or clusters [Fig. 2] among the possible output labels.

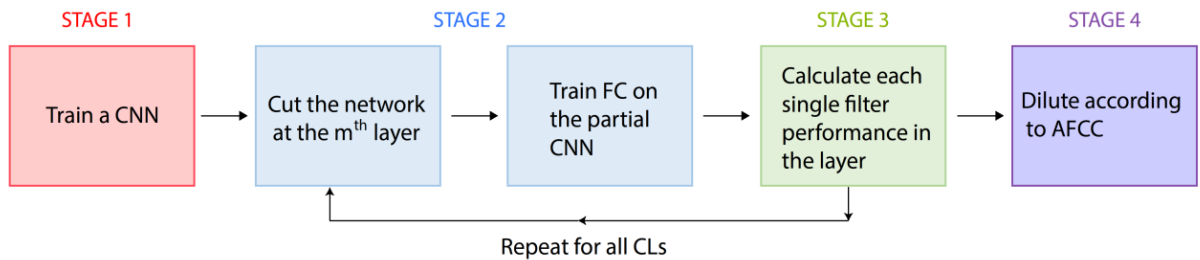

FIG. 1. **Flowchart for calculating single filter performance.** A deep CNN is trained to minimize the loss function (Stage 1). The CNN is then cut at the $m$th layer which is FC to the output and trained to minimize the loss with fixed weights of the previous $m$ layers (Stage 2). The properties of a specific filter are calculated by silencing all the weights, except those emerging from that specific filter. The matrix elements representing the average output fields on each output unit for a specific input label are calculated using the training dataset. The clusters and noise elements of each filter are then calculated using the matrix elements, in each layer (Stage 3). Finally, learning using a diluted deep architecture in accordance with the calculated clusters, obtained by the applied filter's cluster connections (AFCC) technique, is performed (Stage 4).

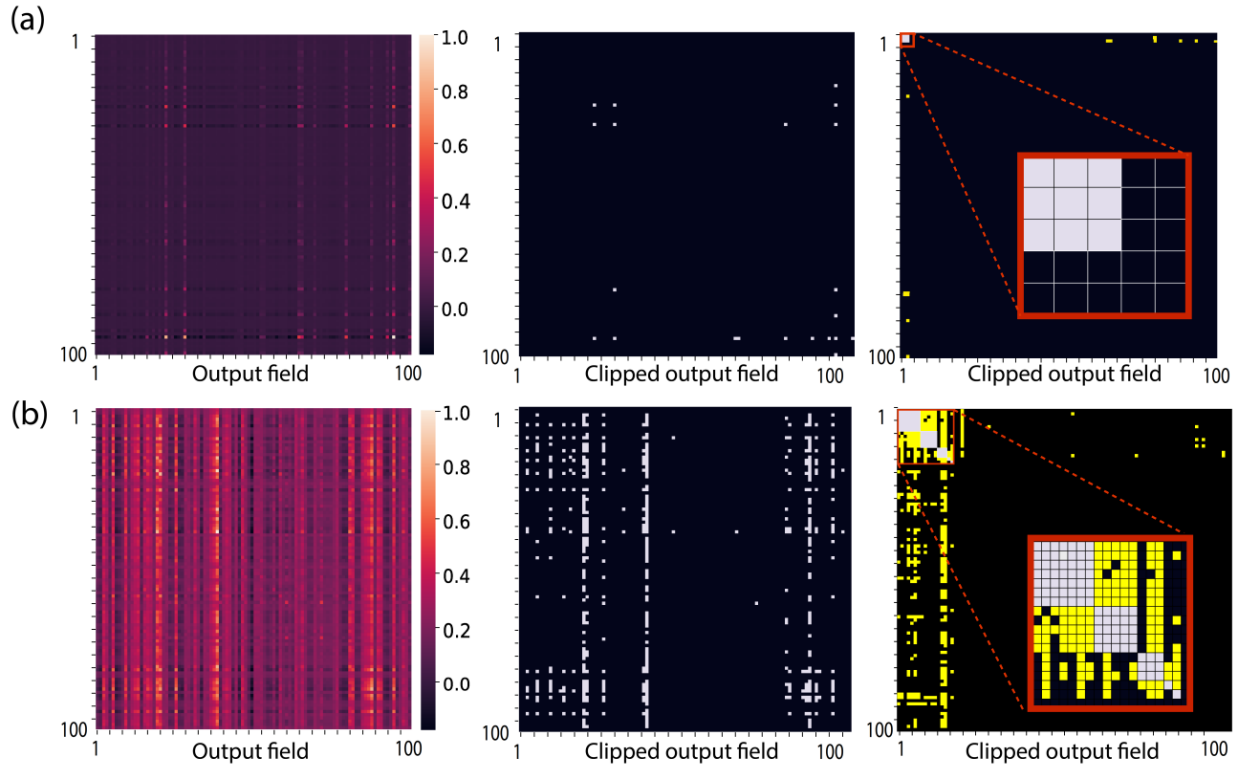

FIG. 2. **Single filter performance.** (a) Left: The matrix element $(i,j)$ of a filter belonging to layer 10 of VGG-11 trained on CIFAR-100. Element $(i,j)$ of the matrix represents the averaged field that was generated by label $i$ test inputs on output $j$, where the matrix elements were normalized by their maximal element. Middle: The Boolean clipped matrix (0/1 is represented by black/white pixels) following a given threshold, $th = 0.3$. Right: Permutations of the clipped matrix labels resulting in a single diagonal cluster of $3 \times 3$ (magnified upper-left corner red box), where above-threshold elements out of the diagonal cluster are noise elements, denoted by yellow. (b) The same images as in (a) but for a filter belonging to layer 7, where the specified filter consists of five clusters of sizes $7, 5, 3, 1, 1$.

This newly discovered mechanism enables the quantification of each filter's signal manifested on average in the small set of labels that it recognizes. The overall macroscopic signal to noise ratio of each layer as well as the entire network can be deduced from the microscopic behavior of each filter and their cooperative functionality. This mechanism is similar to a statistical mechanics viewpoint for analyzing macroscopic behavior of complex systems stemming from of the cooperative behavior of its microscopic entities. The quantifiable language presented by the mechanism also enables a quantification of the signal to noise ratio, which increases along the layers, indicating an accumulating signal and a relatively diminished noise resulting in enhanced accuracy towards the output layer. This work exemplifies how this newly discovered mechanism underlying successful deep learning can be harnessed to enable the system's global decision pruning technique and single filter performance to prune individual filters in each layer. It enables massive pruning that is aimed at preserving each layer's signal and the network's overall accuracy. This type of pruning method is time independent which is similar to disordered magnetic systems with quenched disorder [18,19].

The remainder of this paper is organized as follows. First, section 2 presents the AFCC pruning. Sections 3, 4, 5 examine the AFCC and compares it to artificial and random AFCC, tested on VGG-11 and EfficientNet-B0. Finally, Sections 6 and 7 present the single node performance and FC pruning and discussions, respectively.

## II. APPLYING FILTER'S CLUSTER CONNECTIONS (AFCC) PRUNING

Previous works demonstrated how quantification of the performance of a single filter can prune the FC layer to the output without affecting the overall accuracy [17]. This study demonstrates how quantification of the single filter performance enables highly pruned CLs, diluting constituting a large fraction of deep architecture parameters [Fig. 1, Stage 4], without affecting its overall accuracy. Specifically, the knowledge of single filter performance enables the dilution of the network's filter connections in each layer by keeping only those connections that link filters with similar cluster labels, AFCC [Fig. 3].

Prior to performing AFCC pruning, all the filters of a specific layer generate an output field on all the filters of the consecutive layer, regardless of their cluster labels [Fig. 3a]. On average, each filter generates a strong output field on its cluster labels, whereas the outside cluster elements generate noise, which is usually negligible[16]. This means that for a specific filter, filters from the previous layer that do not share similar cluster labels serve no purpose in generating the signal of that filter. By connecting each filter to the filters of the next layer that share at least one common label in their cluster [Fig. 3b], unwanted noise can be minimized from previous filters that respond strongly to different labels. Consequently, AFCC significantly reduces the number of required connections.

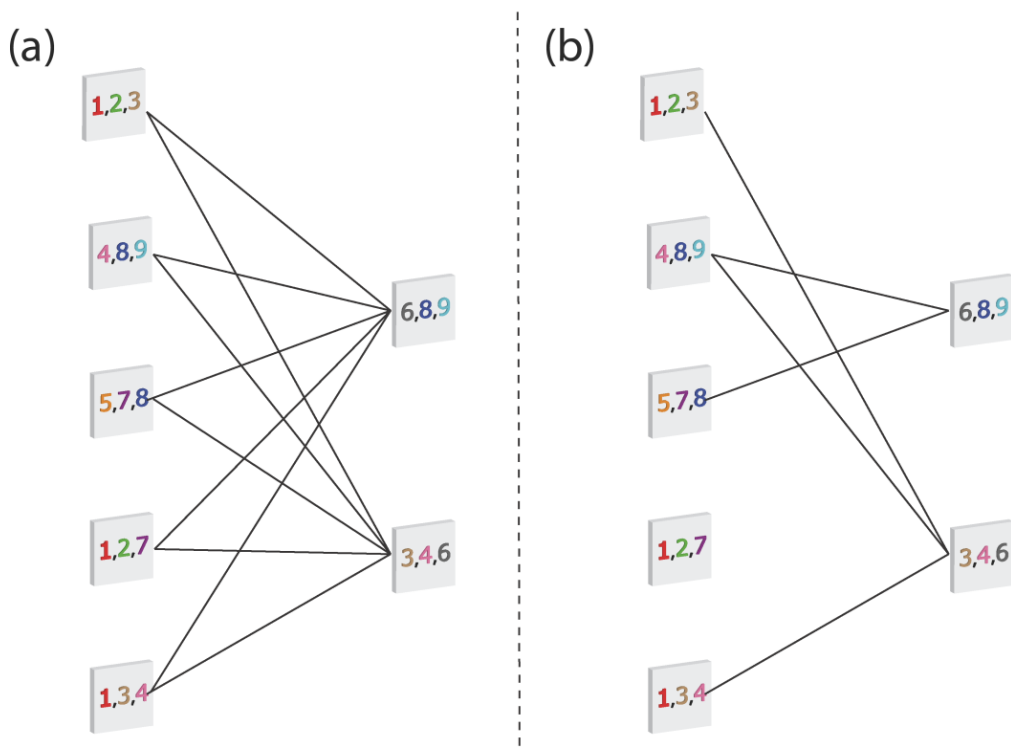

FIG. 3. **Pruning by AFCC.** (a) Scheme of filters' clusters of two consecutive layers of a CNN, before AFCC pruning, where each filter (denoted by gray squares) is connected to all filters in the consecutive layer. The colored numbers represent each filter's cluster labels. (b) The same scheme after pruning, where filters are connected only to filters who share at least one cluster label.

This AFCC method significantly reduces the required filter connections between CLs, conserving only signal-generating connections. This technique requires prior knowledge of the layers' clusters and must be performed after a training session.

An additional suggested pruning scheme, the artificial AFCC (A-AFCC), is artificially associated in advance with each filter with a set of cluster labels. This method skips the second stage, where each of the layers is FC to the output [Fig. 1]. It uses prior knowledge of the statistics of the number of above-threshold diagonal elements, the average cluster size times the average number of clusters in each layer, to dilute the connectivity between filters while training the entire network in the first stage only. This method results in a network trained in an already pruned manner and significantly reduces the complexity associated with the necessity of training the network twice: first to determine the cluster of each filter and then prune the CNN using AFCC followed by fine-tuning the pruned network.

Finally, random AFCC (R-AFCC) is proposed, where the dilution of weights is randomly selected according to the connectivity probability rates based on AFCC pruning statistics. Therefore, diluting a similar number of connections as in the AFCC without considering each filter's cluster.

The proposed pruning methods are examined and compared on VGG-11 and EfficientNet-B0[20], both of which are trained on the CIFAR-100 dataset[21].

### III. AFCC ON VGG-11 AND EfficientNet-B0

The AFCC method is exemplified here on VGG-11, an architecture similar to VGG-16[22] without the last convolutional layers: $11-13$, and with only one FC layer between the last $10^{th}$ CL and the output layer. VGG-11 achieved accuracies similar to those of VGG-16 on the CIFAR-100 dataset. The redundancy of the last three convolutional layers can be attributed to the receptive field covering the entire input size in layer $10$. The three CLs $(3 \times 3)$, layers $8-10$, generate a $7 \times 7$ receptive field[23] covering an image size of $4 \times 4$. Hence, layers $11-13$ are redundant for small images such as those with an input size $32 \times 32$.

The training of VGG-11 on CIFAR-100 [Fig. 2] with the optimized parameters yielded a test accuracy of approximately $0.75$ [Table I], which was slightly higher than the previously obtained accuracy[24]. Next, the weights of the trained layers were held unchanged, and the outputs of layers $5$ through $10$ were FC with random initial weights to the output layer. For every $m^{th}(\geq 5)$ layer, a FC layer was then trained to achieve minimal loss while maintaining the weights of the CLs fixed. After training of the FC statistical features such as cluster size and noise were obtained [Table I].

The pruning of each layer is performed in accordance with its clusters, where the signal of each filter is maximal for inputs belonging to its clusters and small for the rest[16]. As the diagonal size typically shrinks monotonously until the last layer, pruning will grow in relative size towards the output layer. After pruning, a short fine-tuning session of a few epochs was performed to enhance accuracy. The AFCC method is displayed here on a VGG-11 architecture, which yielded a test accuracy of $0.76$ [Table I], which is slightly higher than that of VGG-16. The results indicate that with AFCC, the overall accuracy of the system remains the same [Table II]. After performing AFCC, the accuracy decreased slightly from $0.76$ to $0.758$, yet after a short training session, it recovers to $0.76$.

| \multicolumn{6}{c}{VGG-11 on CIFAR-100 post AFCC pruning} |
|---|---|---|---|---|---|
| Layer | Accuracy | $N_c$ | $C_s$ | Diagonal | Dilution Rate |
| 11 | 0.76 | - | - | - | 0.95 |
| 10 | 0.76 | 1.53 | 3.01 | 4.61 | 0.90 |
| 9 | 0.76 | 1.12 | 2.19 | 2.45 | 0.90 |
| 8 | 0.75 | 2.17 | 2.23 | 4.83 | 0.74 |
| 7 | 0.71 | 2.64 | 2.58 | 6.81 | 0.64 |
| 6 | 0.65 | 3.54 | 2.46 | 8.71 | 0.50 |
| 5 | 0.60 | 4.45 | 2.58 | 11.48 | - |

TABLE I. Accuracy and expected dilution rate per layer for layers 5–11 and their statistical features of VGG-11 trained on CIFAR-100. Accuracy: Accuracy of the layer. $N_c$: Average number of clusters per filter. $C_s$: The average cluster size per filter. Diagonal: The representative diagonal size: $N_c \times C_s$. Dilution Rate: The probability of being diluted (pruned) based on the extracted statistical features. Layer 11 is a fully connected layer and thus has no filters. Layer 5 was not diluted, but the layer's information is needed to perform the pruning of the weights between layers 5 and 6.

Performing AFCC pruning on layers lower than layer 5 resulted in slightly lower accuracies in the output of layer 11. This can be attributed to the high percentage of negative average fields generated by each filter or the fact that clusters in lower layers are considerably noisier than those in higher layers, resulting in a less efficient pruning process[25].

Similar results were obtained for EfficientNet-B0 on the CIFAR-100. The EfficientNet-B0 pretrained on ImageNet was used for transfer learning[26] of CIFAR-100. Optimal parameters resulted in an accuracy of $0.87$ on the validation set.

After training, the clusters of the layers were obtained using the same process as for VGG-11, and AFCC was applied to the upper layers of EfficientNet-B0, specifically, the FC layer and the two uppermost CLs, in the eighth stage of EfficientNet-B0.

| EfficientNet-B0 on CIFAR-100 post AFCC pruning | | | | | |
|---|---|---|---|---|---|
| Layer | Accuracy | $n$ | $N_c$ | $C_s$ | Diluted Rate |
| Stage 9 | 0.87 | 35.92 | 1.33 | 3.11 | 0.96 |
| Stage 8 block 3 | 0.86 | 757.41 | 3.71 | 2.16 | 0.74 |
| Stage 8 block 2 | 0.85 | 118.95 | 1.42 | 3.44 | 0.71 |

TABLE II. Accuracy and silenced rate per layer and their statistical features of EfficientNet-B0 on CIFAR-100. Accuracy: Accuracy of the layer. $n$: Average filter noise. $N_c$: Average number of clusters per filter. $C_s$: Average cluster size per filter. Diluted Rate: The probability of being diluted (pruned).

After performing AFCC for the three uppermost convolutional layers, together with a short training session of a few epochs, an accuracy of $0.87$ is recovered, indicating that pruning using AFCC does not affect the overall accuracy.

## IV. ARTIFICIAL AFCC ON VGG-11

Artificial AFCC (A-AFCC) capitalizes on the nature of cluster formation in filters and aims to create an already pruned system during the training phase. Using the already acquired knowledge of the number of above-threshold diagonal elements, the A-AFCC attempts to train the architecture while imitating this feature.

Before training, each filter is artificially assigned a set of labels representing the labels that appear in its clusters, where each label appears evenly across the clusters of the layer. Next, AFCC is performed on the filters in consecutive CLs that artificially include at least one common label. This creates AFCC pruning prior to the training phase, and allows training to be performed while the architecture is already pruned.

The optimal results were achieved when the cluster size was chosen to be larger than the average diagonal received on VGG-11 when trained on CIFAR-100 ($diagonal = cluster\ size \times num\ clusters$). A single large cluster was selected for each filter because a combination of small clusters could not be artificially created. The single large cluster for each filter allows it to break apart in the learning process into smaller clusters with more than one cluster per filter.

For the A-AFCC of VGG-11 on CIFAR-100, the best results were achieved when an artificial cluster size of 7 was selected for the $10^{th}$ layer, and the cluster size increased in increments of $+1$ until it reached layer 5, granting it a cluster size of 12 [Table I]. Training the system in this manner yielded results that were very close to those of the original training method of $0.757$ which was essentially equivalent to $0.76$. The accuracy decrease from layer 5 to layer 6 is attributed to the dilution that begins with the weights between layers 5 and 6.

| VGG-11 on CIFAR-100 using A-AFCC | | | | |
|---|---|---|---|---|
| Layer | Accuracy | Artificial $N_c$ | Artificial $C_s$ | Diluted Rate |
| 10 | 0.76 | 1 | 7 | 0.93 |
| 9 | 0.23 | 1 | 8 | 0.55 |
| 8 | 0.31 | 1 | 9 | 0.46 |
| 7 | 0.21 | 1 | 10 | 0.37 |
| 6 | 0.19 | 1 | 11 | 0.29 |
| 5 | 0.45 | 1 | 12 | 0.23 |

TABLE III. A-AFCC accuracy per layer and artificial cluster size of VGG-11 on CIFAR-100. Accuracy: Accuracy of the layer. Artificial $N_c$: denotes the number of clusters per filter expected from the A-AFCC. Artificial $C_s$: the artificial cluster size used in the AFCC, the last layer (layer 10) is given an artificial expected cluster size of 7 and is incremented by 1 every subsequent layer below. Diluted Rate: The probability of being diluted (pruned).

When performing A-AFCC on layers lower than layer 5 a considerably higher pruning was achieved by diluting the lower layers; however, the accuracy was slightly decreased from $0.76$ to $0.75$. The artificial clusters up to layer 2 were similarly chosen, with an initial cluster size of 7 for the $10^{th}$ layer and with increments of 1, resulting in a single cluster of size 15 on layer 2. Layer 1 cannot have an artificial cluster size because it is an input layer.

**V. RANDOM AFCC ON VGG-11**

Another pruning method involves random silencing of each layer, regardless of the clusters in each filter. Two methods exist in this scenario, both of which were tested on VGG-11 trained on CIFAR-100, which reached an optimal accuracy of 0.76. The first method, pruning entire filter connections according to the percentage of expected silenced elements as calculated from the cluster size of each filter, regardless of the cluster elements they possess.

Randomly pruning entire filter connections in accordance with the expected pruning percentage considerably lowered the accuracy to 0.01, or a mere guess made by the network on the 100 labels. After a short training session, the system reached an accuracy of 0.61, a stark testament to the damage to the accuracy owing to its deviation from the AFCC mechanism. Random pruning does not conserve the dynamics of the cluster labels of each filter; it connects a single filter to a different combination of cluster labels in the preceding and succeeding layers. The random connections also prevent the system from generating new clusters to match the new connections because of their random, non-dynamic, preserving nature. Notably, at higher levels, a silencing ratio of 95% is expected [see Table I], and such an unmitigated pruning technique will considerably affect the accuracy if ill-planned and completely disregards the flow of information.

The second random technique involves pruning according to the percentage of expected silenced elements across filters, in which each individual weight element is diluted with a probability according to the percentage of expected diluted elements, as shown in Table I.

This pruning technique yields similar results. After pruning, the network that yielded an accuracy of 0.76 drops to 0.01. Results of the post-short training session were slightly more favorable and reached an accuracy of 0.68. This can be because the important connections affiliated with the clusters of each filter partially remained, owing to the random pruning of each individual weight.

## VI. SINGLE NODE PERFORMANCE AND FC PRUNING

The AFCC capitalizes on performing massive pruning based on the system's global decision and the classification task on which it trained upon. This global pruning technique was demonstrated in this study between a pair of CLs; however, it can also be applied to a pair of FC layers, where a layer's decision is decided upon by the node's signal strength for different input labels.

Similar to the AFCC, the FC output units are connected to the output layer using a new FC layer, and all remaining layers remain fixed during the second training phase. After minimal loss is achieved by training the new FC layer, the single node performance is determined using a similar procedure used to determine the single filter performance [Fig. 2]. A threshold is used to distinguish between matrix elements that belong to the signal of the node, and clusters and noise elements are identified using the permutation of the matrix. Subsequently, pruning was performed similarly, where each node omitted its output only to nodes that shared at least one similar cluster label.

The results are presented on VGG-16, which comprises $13$ CLs and $3$ FC layers, trained on CIFAR-100 and achieved an accuracy of $0.75$. Pruning was performed between the second and third FC layers, each consisting of $4096$ nodes, connected by ~16M weights that constitute $50\%$ of the total VGG-16 weights. A high threshold of $0.98$ was used to deduce the Boolean matrix elements, resulting in an average cluster size of $1$ and a single cluster for each node with negligible noise elements, $\ll 1$. A threshold of $0.98$ results in a pruning rate of $99\%$, reducing the 16M interconnected weights of the FC layers to 167K, without reducing the overall accuracy.

Furthermore, it was discovered that many nodes had no above-threshold diagonal matrix elements, indicating that they omitted noise rather than a genuine signal. Pruning these nodes completely did not affect the accuracy and served as further pruning. In the first FC layer, approximately $600$ nodes out of the $4096$ are solely noise nodes, and in the second FC layer, approximately $1000$ nodes out of the $4096$ have no above-threshold diagonal elements. Preliminary results indicated that high pruning of weights between three consecutive FC layers of VGG-16, between layers $13$ and $14$, and between layers $14$ and $15$ achieved an accuracy of $0.75$.

Using a lower threshold such as $0.6$ resulted in larger clusters and more noise [Fig. 4], which resulted in a lower pruning rate. A threshold of $0.98$ enables high pruning without affecting the overall accuracy.

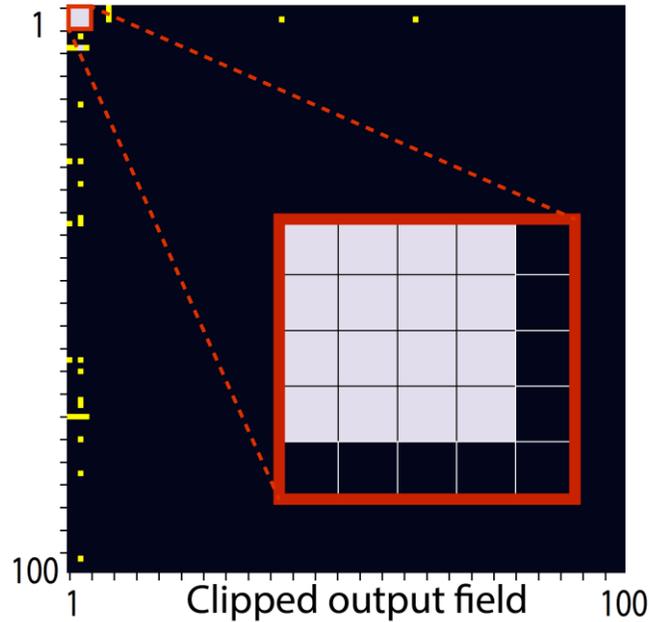

FIG. 4. **Single node performance of FC layer.** The Boolean clipped permuted matrix element $(i,j)$ of a filter belonging to layer $15$ of VGG-16 trained on CIFAR-100. Element $(i,j)$ of the matrix represents the clipped above a threshold of $0.6$ averaged field that was generated by label $i$ test inputs on an output $j$, and permuted similar to the right panel in [Fig. 2].

**VII. DISCUSSION**

The discovery of the underlying mechanism of deep learning, which was quantitatively examined for VGG-11 and EfficientNet-B0 architectures trained on the CIFAR-100 and ImageNet datasets, paved the path for a more refined understanding of how deep learning networks actually work.

This better understanding facilitates a more effective pruning method of the AFCC, exploiting the fact that each filter sends out a signal in the form of a small labels' cluster and emits, on average, a very small field for non-cluster input labels. This enables a very high and effective pruning technique that minimizes each layer's connections to filters with at least one similar cluster label, rather than contemporary pruning techniques that primarily rely on random dilution processes[27-30].

It is important to note that while AFCC pruning is capable of conserving the overall accuracy, it does not increase it. This can be attributed to the fact that the output gap in both correct and incorrect test output predictions is already large and thus the overall decision is not affected by the changes to the signal or noise elements.

The presented underlying mechanism for successful deep DL leads to an efficient quenched pruning method which is reminiscent percolation with the following difference: In percolation the existing links between nodes are known in advanced based on their spatial proximity or mutual activity [31], while in the proposed method the dominating links between nodes are not given in advance but rather discovered through the training process. This quenched pruning is also within the line of biological neural networks which are highly diluted [32-34]. This stands in contrast to annealed pruning methods such as dropout [35] which are common in the computer science community.

This enormous pruning can exceed 90% in certain layers, significantly reducing the number of required parameters in the network ($PARAMs$) and floating-point operations per second ($FLOPs$) needed for the feedforward and backpropagation calculation steps in each iteration. The reduction in backpropagation could be more effective because a large dilution alleviates the numerous calculation steps required from one layer to another. Specifically, the reduction in $PARAMs$ of each CL is given by

$kernel\ size \times input\ channels \times output\ channels \times survival\ rate$,

and the reduction in FLOPs is given by

$input\ size \times kernel\ size \times output\ channels \times survival\ rate$,

where $survival\ rate = 1 - diluted\ rate$. For VGG-11 out of 16, the AFCC reduced the needed $0.286$ G-Macs (Giga Multiply-Accumulate Operations) to $0.197$ G-Macs, a reduction of ~31% [Table I]. For the A-AFCC, the number of operations was reduced to $0.222$ G-Macs, a reduction of ~22% [Table III].

Further enhancing the AFCC method to the A-AFCC, which performs pruning of the architecture prior to the training phase, capitalizing on the nature of the cluster mechanism and artificially assigning cluster elements to each filter and connecting the filters accordingly, achieved similar accuracies of up to $0.01$ difference. Although this technique uses the already known statistics of each layer on an already regularly trained architecture to achieve optimal results, we can guess the required cluster sizes for the A-AFCC, and it is robust to a large variety of different guesses. Using the comprehension of shrinking cluster sizes up to the output layer and assigning a larger cluster size as a preemptive measure will be more than sufficient to ensure optimal system performance. In general, iterating over different statistical features for the A-AFCC will result in a pruned system that requires less computation than the AFCC method; however, further research is required on this subject.

One limitation of the A-AFCC technique is that artificial clusters can only be created as a single large cluster for each filter constituting all its diagonal elements rather than a combination of individual smaller clusters. This limitation hinders the artificial creation of clusters and forces a lower pruning rate or subsequently, requires the trained network's extrapolated statistical features to perform higher pruning without affecting overall accuracy. The ability to artificially create cluster combinations within each filter is a prospect for further research.

Performing this enormous pruning randomly by eliminating the expected pruned percentages of the connections of each filter to the subsequent layer significantly reduces the accuracy of the overall architecture, adhering to the requirement of maintaining essential signal connections. Although randomly pruning the weight elements in the filter can achieve higher accuracy with similar pruning rates, it is still far below the accuracy obtained by the AFCC. Notably, such random pruning does not reduce the number of $FLOPs$ because each filter will still generate a field to all filters in the subsequent layer, thereby requiring more operations than the usual AFCC method. Randomly pruning entire filters in a manner similar to AFCC will result in lower $FLOPs$; however, this will considerably affect the accuracy of the architecture, thereby being a counterproductive pruning technique.

AFCC global pruning technique can effectively reduce the memory and energy usage of trained architectures, making it the forefront of hardware optimization for phones and other electrical devices that rely heavily on small and effective components.

It is important to compare the AFCC method with other existing dilution techniques. One of the most common pruning techniques is dropout [35], an annealed pruning technique which randomly time-dependently dilutes the nodal outputs at a specified layer during the training process. The key difference stems from the fact that AFCC, a global decision pruning technique, enables pruning weight connections that are less fundamental to the architecture's performance. Other quenched dilution techniques such as Low-rank decomposition of weights [36] can be compared as well. This method leverages tensor decomposition techniques to perform filter approximations and thereby reducing the needed computation and memory to perform a given task. However, it is a local pruning technique that does not take into account the global decision of the labels' classification by the entire network. Comparing the Low-ranking decomposition and the global AFCC pruning techniques without affecting the accuracy is beyond the scope of this research and deserves further work.

Notably, the proposed pruning technique is expected to be more efficient for parallel computation units assigned to each node of the architecture. Although the number of layers remain the same, the number of required operations in each calculation step of each layer is greatly reduced. For sequential computation, the number of operations remains similar because of the nature of matrix multiplication that modern-day programs use to implement deep learning architectures, although it greatly benefits from zeroing many components. However, the extensive improvement this technique can have in memory usage remains the same because the $PARAMs$ variable is significantly reduced.

The core essence of the AFCC technique, which performs pruning based on the final global classification result rather than using local phenomena for each individual variable, can be extended to other networks and architectures consisting of FC layers. Pruning an FC layer based on the single node performance, proved to be extremely efficient, diluting nearly $99\%$ of the parameters in the FC layer without any reduction in the overall accuracy. FC pruning can be extremely beneficial for architectures with large consecutive FC layers such as transformers[37]. These results demonstrate the universality of pruning in accordance with global decisions and further extrapolation to other layer types is required.

An important aspect of the AFCC is that it is a quenched pruning technique that is independent of the input presented at a given time. Further enhanced pruning to this quenched pruning can be achieved using an annealed pruning based on the layers' outputs which are zeroed owing to the ReLU activation function[34]. For instance, for VGG-11 with quenched AFCC, the 9th layer survival rate of each weight is $0.1$, [Table I] and the layer's annealed ReLU outputs are $82\%$ zero. Hence, a potential dynamic pruning of $0.1 \times 0.18 = 0.018$ is suggested, thereby potentially reducing the layer output units from $8,192$ to $139$ nodes. This large pruning potential requires further research, which is beyond the scope of this study.

**APPENDIX A: MATERIALS AND METHODS**

*1. Architectures and training of the fully-connected layers*

In this study, two architectures were examined: VGG-11, a modification of VGG-16[22] constituting only $10$ CLs instead of $13$; and EfficientNet-B0[20]. The architectures were trained to classify the CIFAR-100 dataset[21], with no biases on the output units to ensure that each filter's effect on the output fields would be exemplified and not overshadowed by the much larger biases. Removing the biases of the output layer did not affect the architectures' average accuracies, in comparison to architectures trained with output biases.

The evaluation was performed by taking an architecture trained on the entire dataset, cutting it at designated layers and training a new FC layer between the output of that specific layer and the output layer. During this training process, only the FC layer was trained, whereas the weights and biases of the rest of the architecture remained fixed.

*2. Dataset and preprocessing*

The image pixel in the CIFAR-100 dataset[21] were normalized to the range $[-1, 1]$ by dividing by $255$ (the maximal pixel value), multiplying by $2$, and subtracting $1$. In all simulations, data augmentation derived from the original images was performed, by random horizontal flipping and translating up to four pixels in each direction.

*3. Optimization*

The cross-entropy cost function was selected for the classification task and minimized using the stochastic gradient descent algorithm[2,38]. The maximal accuracy was determined by searching through the hyper-parameters (see below). Cross-validation was performed using several validation databases, each consisting of a randomly selected fifth of the training set examples. The average results were within the same standard deviation (Std) as the reported average success rates. The Nesterov momentum[39] and L2 regularization method[40] were applied.

*4. Hyper-parameters*

The hyper-parameters $\eta$ (learning rate), $\mu$ (momentum constant[39]), and $\alpha$ (L2 regularization[40]) were optimized for offline learning, using a mini-batch size of 100 inputs. The learning-rate decay schedule[41] was also optimized. A linear scheduler was applied such that it was multiplied by the decay factor, q, every $\Delta t$ epochs, and is denoted below as $(q, \Delta t)$. Different hyper-parameters were used for each architecture.

*5. VGG-16 hyper-parameters*

VGG-16 was trained over 300 epochs using the following hyper-parameters to achieve maximal accuracy on CIFAR-100:

**TABLE V.** Hyper-parameters for VGG-16 trained on CIFAR-100.

| VGG-16 on CIFAR-100 | | |
| --- | --- | --- |
| $\eta$ | $\mu$ | $\alpha$ |
| 5e-3 | 0.93 | 1.5e-3 |

The decay schedule for the learning rate during training of the entire system is defined as follows:

$$(q, \Delta t) = (0.6, 20)$$

For the training of the FC layer, $\eta = 0.01$, $\mu = 0.975$, $\alpha = 1e-3$, with a learning rate scheduler of $q = 0.975$ every 1 epoch, and the other weight values and biases of the architecture remained fixed.

*6. VGG-11 hyper-parameters*

VGG-11 was trained over 300 epochs using the following hyper-parameters to achieve maximal accuracy on CIFAR-100:

**TABLE VI.** Hyper-parameters for VGG-11 trained on CIFAR-100.

| VGG-11 on CIFAR-100 | | |
|---|---|---|
| $\eta$ | $\mu$ | $\alpha$ |
| 5e-3 | 0.93 | 1.5e-3 |

The decay schedule for the learning rate during training of the entire system is defined as follows:

$$(q, \Delta t) = \begin{cases} (0.65, 20) & 160 > epoch \\ (0.55, 20) & 160 \leq epoch \end{cases}$$

For the training of the FC layer, $\eta = 0.01$, $\mu = 0.975$, $\alpha = 1e-3$, with a learning rate scheduler of $q = 0.975$ every 1 epoch, and the other weight values and biases of the architecture remained fixed.

*7. EfficientNet-B0 hyper-parameters*

The pretrained EfficientNet-B0 on ImageNet was loaded and was then trained over 300 epochs using the following hyper-parameters to achieve maximal accuracy on CIFAR-100:

**TABLE VII.** Hyper-parameters for EfficientNet-B0 trained on CIFAR-100.

| EfficientNet-B0 on CIFAR-100 | | |
|---|---|---|
| $\eta$ | $\mu$ | $\alpha$ |
| 1e-2 | 0.9 | 1e-3 |

The decay schedule for the learning rate during training of the entire system is defined as follows:

$$(q, \Delta t) = \begin{cases} (0.65, 20) & 160 > epoch \\ (0.55, 20) & 160 \leq epoch \end{cases}$$

For the training of the FC layer, $\eta = 0.004$, $\mu = 0.975$, $\alpha = 1.5e-3$, with a learning rate scheduler of $q = 0.65$ every 20 epochs, and the other weight values and biases remained fixed.

## 8. Calculation of clusters and noise

For CIFAR-100, the 100 output fields of each filter were summed over all 10,000 inputs of the test set, resulting in a $100 \times 100$ matrix, where each cell $(i,j)$ represents the summed field of output field $j$ for all test set inputs of label $i$. The matrix was normalized by dividing it by its maximal value, resulting in each matrix having a maximum value of 1. The clipped Boolean output field matrix was calculated, where each element whose value was above a threshold $th = 0.3$ was set to 1 and all others were zeroed. Similar qualitative results were obtained for different thresholds $[0.1, 0.6]$, where 0.3 represents the matrix's distribution cutoff between high and low values.

The axes were then permuted such that all labels belonging to a cluster were grouped consecutively, thereby displaying the clusters in an adjacent fashion, where they were shown as a diagonal block of elements with value 1 [Figure 2]. Each cluster was defined as a subset of $n$ indices, where for each $i, j \in n$ elements $(i,j)$ have the value of 1. The size of the cluster was defined as $n$, where $n$ is the number of labels whose all pairs (and their permutation) are equal to 1, thereby forming a cluster. The minimal cluster size is 1, that is one element on the diagonal, or 100, the entire matrix. The elements equal to 1 were then colored white, representing that they belong to a cluster in the filter, while non-cluster cells with a value of 1 were classified as above-threshold external noise and were colored yellow; cells with a value of $-1$ were colored green and the rest were zero and colored black.

The clusters were calculated by running along the diagonal from index (0,0) to (99,99), where the first $(i,i)$ element with a value of 1 was initially designated as a cluster of size $1 \times 1$. The next $(j,j)$ element, where $j \neq i$, with a value of 1 was then checked to determine if it can complete a cluster with $(i,i)$; if yes, it was added to the cluster and the next diagonal element with a value of 1 was checked. This process was repeated for all value 1 cells in the diagonal, as long as there were elements that did not belong to a cluster. This process is not uniquely defined, that is, the order by which the indices are iterated can change the outcome of the clustering process. For example, a filter with two clusters of sizes $3 \times 3$ and $1 \times 1$ retrieved by iterating from $0$ to $99$ can yield, in some very rare scenarios, two clusters of size $2 \times 2$. While possibly alternating the results of a single filter, the overall obtained averaged results remain the same when performing cluster creation while iterating in reverse order, because these scenarios are very rare and occur in a negligible number of filters.

**APPENDIX B: STATISTICS, HARDWARE AND SOFTWARE**

*1. Statistics*

Statistics for all results were obtained using at least five samples.

*2. Hardware and software*

We used Google Colab Pro and the available GPUs. We used PyTorch for all the programming processes.